\documentclass[aip,graphicx,amsmath,amssymb,preprint]{revtex4-1}
\usepackage{CJKutf8}
\usepackage{lineno,hyperref}
\usepackage[normalem]{ulem}
\usepackage{graphics}
\usepackage{amsmath,bm}
\usepackage{amssymb}
\usepackage{graphicx}
\usepackage{makecell}
\usepackage{bm}
\usepackage{amsfonts}
\usepackage{color}
\usepackage{epstopdf}
\usepackage{booktabs}
\usepackage{array}
\usepackage{hyperref}
\usepackage{graphics}
\usepackage{epsfig}
\usepackage{bm}
\usepackage{epic,eepic}
\usepackage{epsfig}
\usepackage{pifont}

\usepackage{subfigure}

\usepackage{xspace}
\usepackage{graphicx,color}
\usepackage{amsmath,amssymb,amsfonts,mathrsfs}
\usepackage{tikz}
\usetikzlibrary{shapes, arrows.meta, positioning, calc, fit}
\usepackage{url}
\usepackage[normalem]{ulem}
\usepackage{booktabs}
\usepackage{float}
\usepackage{algorithm} 
\usepackage{mathtools}
\usepackage{color} 
\usepackage{algpseudocode} 
\usepackage{xcolor}
\usepackage[toc,page]{appendix}
\usepackage[capitalize]{cleveref}
\usepackage{mathrsfs} 
\usepackage{algorithm} 
\usepackage{mathtools}
\usepackage{color} 
\usepackage{algpseudocode} 
\usepackage[a4paper, total={7in, 10in}]{geometry}
\usepackage[export]{adjustbox} 
\newcolumntype{P}[1]{>{\centering\arraybackslash}p{#1}}

\makeatletter
\newcommand{\defeq}{\mathrel{\aban@defeq}}
\newcommand{\aban@defeq}{%
  \vbox{\offinterlineskip\check@mathfonts
    \ialign{\hfil##\hfil\cr
      \fontsize{\ssf@size}{\z@}\normalfont def\cr
      \noalign{\kern1\p@}
      $\m@th=$\cr
      \noalign{\kern-.5\fontdimen22\textfont2}
    }%
  }%
}
\makeatother

\newcommand{\bx}{\mathbf{x}}

\newcommand{\bc}{\mathbf{c}}
\newcommand{\bu}{\mathbf{u}}

\definecolor{darkgreen}{rgb}{0.0, 0.5, 0.0}

\newcommand\RE[1]{\textcolor{black}{#1}}

\begin{document}

\title{Fast-Forward Lattice Boltzmann: Learning Kinetic Behaviour with Physics-Informed Neural Operators}

\author{Xiao Xue\footnote{Email: \texttt{x.xue@ucl.ac.uk}}}
\affiliation{ 
Centre for Computational Science, University College London, UK
}
\author{Marco F.P. ten Eikelder}
\affiliation{Institute for Mechanics, Computational Mechanics Group, Technical University of Darmstadt, Germany}
\author{Mingyang Gao}
\affiliation{ 
Centre for Computational Science, University College London, UK
}
\author{Xiaoyuan Cheng}
\affiliation{ 
Centre for Dynamic Systems, University College London, UK
}

\author{Yiming Yang}
\affiliation{ 
Department of Statistical Science, University College London, UK
}
\author{Yi He}
\affiliation{ 
Centre for Dynamic Systems, University College London, UK
}
\author{Shuo Wang}
\affiliation{ 
Department of Physics, Eindhoven University of Technology, Eindhoven, Netherlands
}
\author{Sibo Cheng}
\affiliation{ 
Centre d'Enseignement et de Recherche en Environnement Atmosphérique (CEREA), \'{E}cole des Ponts and EDF R\&D, Institut Polytechnique de Paris, \^Ile-de-France, France
}

\author{Yukun Hu\footnote{Email: \texttt{yukun.hu@ucl.ac.uk}}}
\affiliation{ 
Centre for Dynamic Systems, University College London, UK
}
\author{Peter V. Coveney\footnote{Email: \texttt{p.v.coveney@ucl.ac.uk}}}
\affiliation{ 
Centre for Computational Science, University College London, UK
}
\affiliation{
Computational Science Laboratory, Institute for Informatics, University of Amsterdam, the Netherlands
}
\affiliation{Centre for Advanced Research Computing, University College London, UK}

\date{\today}
\begin{abstract}






The lattice Boltzmann equation (LBE), rooted in kinetic theory, provides a powerful framework for capturing complex flow behaviour by describing the evolution of single-particle distribution functions (PDFs). Despite its success, solving the LBE numerically remains computationally intensive due to strict time-step restrictions imposed by collision kernels. Here, we introduce a physics-informed neural operator framework for the LBE that enables prediction over large time horizons without step-by-step integration, effectively bypassing the need to explicitly solve the collision kernel. We incorporate intrinsic moment-matching constraints of the LBE, along with global equivariance of the full distribution field, enabling the model to capture the complex dynamics of the underlying kinetic system.
Our framework is discretization-invariant, enabling models trained on coarse lattices to generalise to finer ones (kinetic super-resolution).
In addition, it is agnostic to the specific form of the underlying collision model, which makes it naturally applicable across different kinetic datasets regardless of the governing dynamics.
Our results demonstrate robustness across complex flow scenarios, including von K\'arm\'an vortex shedding, ligament breakup, and bubble adhesion.
This establishes a new data-driven pathway for modelling kinetic systems.




\end{abstract}

\keywords{Lattice Boltzmann Equation, Physics-Informed Machine Learning, Neural Operator, Equivariant representation}

\maketitle
\section{INTRODUCTION}
Mathematical models based on partial differential equations are essential tools for simulating dynamical systems across the sciences~\cite{van2009plasma,porte2011large,mehta2014large,kurth2023fourcastnet}. While these macroscopic models underpin much of our understanding of fluid dynamics, plasma physics, and climate science, they often struggle to model microscopic interactions, which are essential for accurately modelling multiphase or multiscale phenomena~\cite{cercignani1969mathematical,lulli2025higher,pelusi2024intermittent}.
Kinetic theory offers a natural framework to bridge microscopic and macroscopic scales by describing the evolution of single-particle distribution functions \cite{kennard1938kinetic,villani2002review,loeb2004kinetic}. Importantly, it plays a central role in addressing Hilbert’s sixth problem, which seeks a rigorous derivation of continuum equations from underlying particle dynamics—a challenge that has recently seen major progress through the work of Deng et al. \cite{deng2025hilbert}. The lattice Boltzmann equation (LBE) has been applied to model a wide range of complex physical systems~\cite{he1997theory, succi2001lattice, lallemand2000theory, lallemand2021lattice,kruger2017lattice}, including multiphase flows~\cite{shan1993lattice,wang2019brief,connington2012review,li2013lattice,luo2021unified,ronsin2022phase,xue2025equivariant,kusumaatmaja2008collapse,sofonea2004finite}, turbulent flows~\cite{boghosian2001entropic,suga2015d3q27,bosch2015entropic,chikatamarla2013entropic}, as well as biomedical applications such as blood flow and hemodynamics in vascular networks~\cite{mazzeo2008hemelb,groen2013analysing,lo2025multi,lo2024uncertainty}. Inspired by Grad's approach~\cite{grad1949kinetic,grad1958principles}, Shan and his coauthors~\cite{shan2006kinetic} have demonstrated a promising systematic method to approximate the collision kernel using higher-order Hermite expansions, which can equivalently represent the Navier-Stokes equations, the Burnett equations, and many others. Beyond its application in hydrodynamics on both the macroscale~\cite{succi2001lattice,hou1994lattice} and microscale~\cite{ladd1993short,xue2018effects,xue2020brownian}, the LB equation also connects to relativistic fluids~\cite{mendoza2010fast,ambrucs2022fast}, thereby extending its relevance to cosmology and plasma physics~\cite{simeoni2024lattice} at the kinetic level. Despite its success, solving the LBE remains computationally intensive.

As a result, new approaches are needed to overcome the computational challenges associated with solving kinetic equations. Machine learning (ML) models, ranging from convolutional neural networks and recurrent architectures to more recent innovations such as transformers and diffusion models, have achieved remarkable success in domains such as computer vision, natural language processing, and generative modelling~\cite{kasneci2023chatgpt,gopalakrishnan2017deep,ding2014data,hou2013model,cheng2025machine}. Motivated by these advances, there is growing interest in applying ML techniques to scientific computing tasks. In particular, ML-based approaches have shown strong promise in fluid mechanics, where they are used for prediction, surrogate modelling, and closure modelling of dynamical systems~\cite{cuomo2022scientific,brunton2020machine,zhong2022low,wen2021towards,ortali2025enhancing}. 
\indent Within fluid mechanics, ML models are typically categorized into two main approaches. The first, referred to as ``hybrid ML'', incorporates machine-learned components into conventional numerical solvers by partially replacing specific terms or submodels in the governing evolution equations~\cite{beck2019deep, duraisamy2019turbulence, bae2022scientific, xue2024physics}. Examples of this approach include the development of turbulent subgrid-scale (SGS)~\cite{beck2019deep, duraisamy2019turbulence, sarghini2003neural, gamahara2017searching, xie2019artificial, mohan2023embedding} and near-wall models~\cite{yang2019predictive, bae2022scientific, xue2024physics}.
The second approach, referred to as ``standalone ML'', aims to replace the entire numerical solution procedure with a data-driven model that directly predicts the system evolution~\cite{li2020fourier}. A prominent example of this approach is the class of neural operators, which aim to learn mappings from input fields to output fields and have been successfully applied to a range of PDE models~\cite{lu2021learning,li2020fourier,kovachki2023neural,cao2024laplace,cheng2025machine,wangxue2025quantum}.

Despite recent progress in scientific ML, its exploration to the LBE remains limited. Most kinetic-focused ML approaches fall under the ``hybrid'' paradigm, owing to the presence of the collision kernel in the LB formulation. These methods typically aim to learn the collision term. For example, Bedrunka et al.\cite{bedrunka2021lettuce} optimised parameters of multiple-relaxation-time (MRT) collision kernels, while Corbetta et al.\cite{corbetta2023toward,ortali2025enhancing} developed a neural network architecture to approximate the discrete Bhatnagar–Gross–Krook (BGK) collision kernel by incorporating mathematical and physical constraints. Extending these ideas, Ortali et al.~\cite{ortali2024kinetic} introduced a three-dimensional neural collision kernel trained on direct numerical simulation (DNS) data to model LB-based subgrid-scale interactions. However, these approaches are typically limited to single-timestep forward predictions and remain constrained by the fine temporal resolution required to resolve collision dynamics.  Consequently, developing methods that bypass time-step constraints and model the forward operator—without explicitly learning the collision kernel—remains an early and largely unexplored area of research. Lat-Net\cite{hennigh2017lat}, for instance, approximates LBE dynamics in a compressed latent space, making it difficult to enforce local physical conservation laws and kinetic constraints, which can lead to degraded accuracy and out-of-distribution scenarios, and limiting its practical applicability in physics-informed modelling.
\setlength{\parskip }{0pt}

\begin{figure}[H]
\textbf{Overview of machine learning methods for kinetic theory}
\vspace{20pt}
\centering

\includegraphics[width=0.95\textwidth]{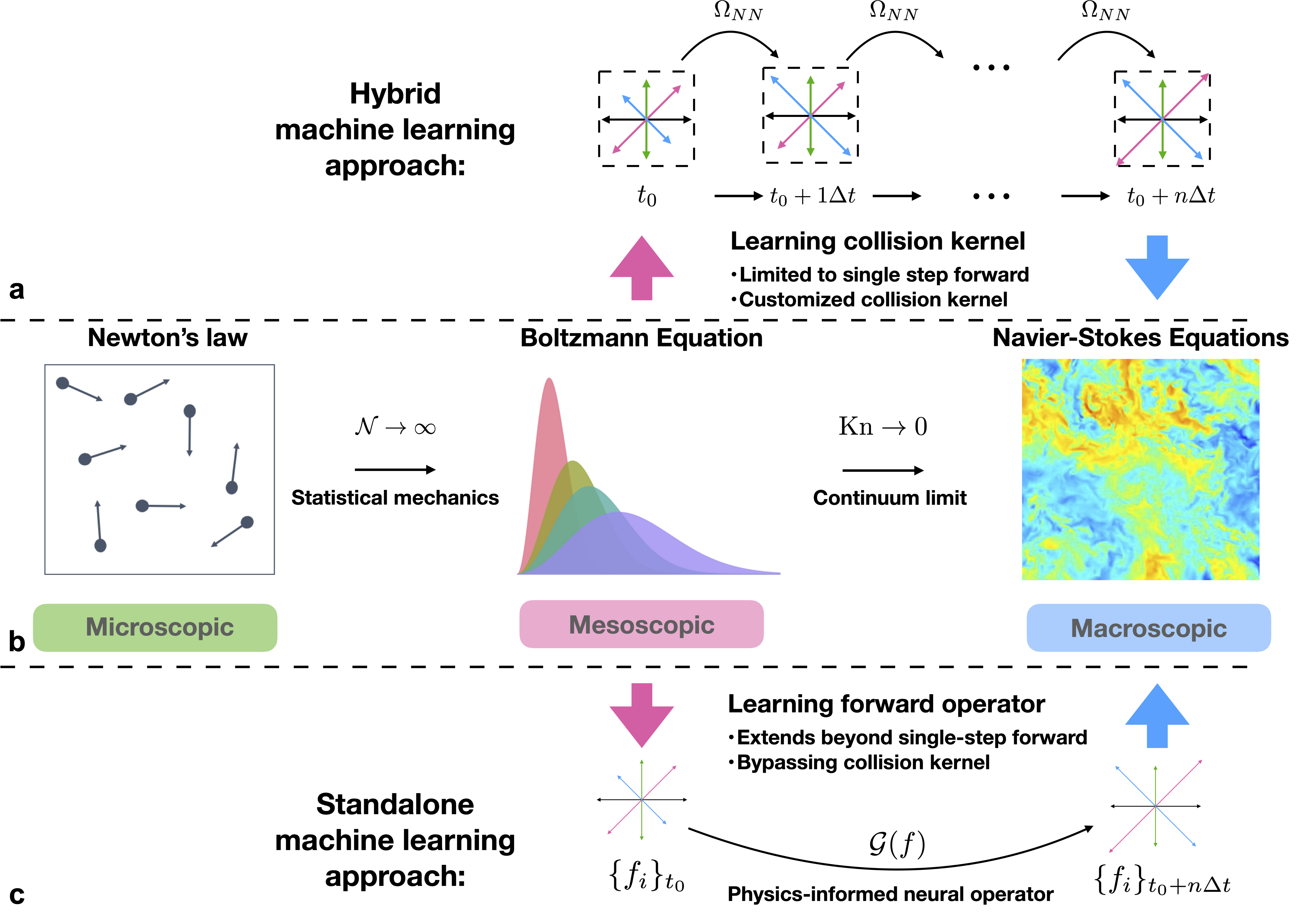}
\caption{A multiscale framework for kinetic theory encompassing microscopic, mesoscopic, and macroscopic perspectives. Panel \textbf{a} outlines the kinetic data-driven hybrid ML approach for kinetic-based methods, which uses neural networks to learn collision kernels limited by the single-step forward for the LB equation. Panel \textbf{b} illustrates molecular dynamics governed by Newton’s laws, focusing on individual molecules. As the molecular count $\mathcal{N}$ grows, statistical mechanics emerge, transitioning to the mesoscopic scale described by the Boltzmann equation. In the continuum limit, where the Knudsen number $\mathrm{Kn} \to 0$, the system converges to the macroscopic hydrodynamic equations. Panel \textbf{c} proposes a novel physics-informed neural operator for LBE, denoted by $\mathcal{G}$, that learns the long-time forward evolution of the single-distribution function fields $\{f_i\}$, enabling large time jumps without explicitly solving the collision kernel.}\label{fig:scale} 
\end{figure}

Here, we propose a physics-informed neural operator (PINO) framework tailored to the lattice Boltzmann equation, enabling fast-forward prediction of single-particle distribution functions while preserving meso-to-macroscale consistency, without relying on explicit learning of the collision kernel. (Fig.~\ref{fig:scale} \textbf{c}). 
To learn physical consistency, we incorporate domain-specific inductive biases into the training process, including local moment-matching constraints and global equivariance of the full distribution function fields, where the global symmetry arises from the underlying local lattice structure. We primarily focus on the U-shaped Neural Operator (UNO) architecture~\cite{rahman2022u} due to its memory efficiency, which makes it well suited for large-scale kinetic simulations. While our main results are based on UNO, ablation studies confirm that the Fourier Neural Operator (FNO)~\cite{li2020fourier} exhibits similar trends under the same physics-informed training regime. These structural constraints enable the model to accurately capture the complex behaviour of the underlying mesoscopic dynamical system while preserving key macroscopic physical properties. We demonstrate that the evolution of single-particle distribution functions can be modeled without explicit dependence on the collision kernel or the underlying lattice structure, while still preserving super-resolution: a model trained at coarse resolution generalizes to finer grids on the same lattice structure, indicating that it captures the underlying kinetic physics rather than memorizing grid-specific patterns. Our approach has been tested on both D2Q9 and D3Q19 lattice models, using a variety of collision operators including the Bhatnagar–Gross–Krook (BGK) model with a Smagorinsky LES closure, the MRT model, and a phase-field formulation of the LBE. This flexibility enables the prediction of complex flow scenarios such as von K\'arm\'an vortex shedding, ligament breakup, and bubble adhesion. The method's flexibility with respect to lattice configurations and collision models highlights its broad applicability across diverse kinetic systems.


\section{Learning lattice Boltzmann equation via neural operator}

The lattice Boltzmann equation can be seen as a velocity-discretised form of the Boltzmann kinetic equation~\cite{succi2001lattice}, where the continuous velocity variable is simplified by a defined finite set of $Q$ discrete velocities $\left\{\mathbf{c}_i\right\}_{i=0}^{Q-1}$. These discrete velocities $\mathbf{c}_i$ are fictitious vectors defined by the lattice and represent the possible directions for particle streaming; they do not correspond to the physical (macroscopic) velocity of the fluid. This leads to a system of $Q$ single-particle distribution functions $f_i=f_i(\bx, t)$, where each $f_i : \mathbb{R}^d \times \mathbb{R} \to \mathbb{R}$ describes the density of particles moving with velocity $\bc_i$ at position $\bx \in \mathbb{R}^d$ being the spatial dimension and time $t \in \mathbb{R}$:
\begin{equation}
\partial_t f_i(\bx, t) + \bc_i \cdot \nabla f_i(\bx, t) = \Omega(\left\{f_j\right\}_{j=0}^{Q-1}(\bx, t))+ F_i(\mathbf{x}, t).
\label{eq:lbe-cont}
\end{equation}
The left-hand side of Eq.~\eqref{eq:lbe-cont} corresponds to the streaming step, which accounts for the propagation of particles, while the first term on the right-hand side represents collisions that redistribute particles among the discrete velocities, where $\Omega$ is the collision operator and the second term accounts for external forcing $F$. 

Conventional numerical solvers for the LBE are constrained by the explicit treatment of the collision operator, which governs local interactions and facilitates information exchange between neighbouring lattice cells. In contrast to the traditional perspective that decomposes the dynamics into streaming and collision steps at the level of individual cells, we adopt a functional perspective. Specifically, we treat each discrete $f_i(\bx, t)$ of the particle distribution function as a spatially structured field over the entire domain.
The goal is to learn a forward operator that predicts the evolution of the full set $\{f_i\}$ over an extended time interval. 
Let $t$ be the initial time and $t = t' + n\Delta t$ the target time. 
When $n$ is large, local interactions become decorrelated, and the evolution becomes increasingly global.
We define a neural operator $\mathcal{G}$ as:
\begin{subequations}\label{eq: NO G}
  \begin{align}
  \mathcal{G}: \mathcal{A} &\rightarrow \mathcal{U}, \\
  \mathcal{A} &= \left\{ \{f_i(\cdot, t)\}_{i=0}^{Q-1} \mid t \in \mathcal{T}_{\text{in}} \right\}, \\
  \mathcal{U} &= \left\{ \{f_i(\cdot, t)\}_{i=0}^{Q-1} \mid t \in \mathcal{T}_{\text{out}} \right\},
  \end{align}
\end{subequations}
where $\mathcal{A}$ represents the collection of distribution functions $\{f_i\}$ across all discrete velocities $i = 0, \dots, Q-1$ over an input time window $\mathcal{T}_{\text{in}} \subset \mathbb{R}$, and $\mathcal{U}$ represents their corresponding evolution over the output time window $\mathcal{T}_{\text{out}} \subset \mathbb{R}$. Each $f_i(\cdot, t)$ is treated as a spatial field belonging to a function space $\Psi$. The global operator $\mathcal{G}$ is trained to map $\mathcal{A}$ to $\mathcal{U}$, encapsulating the combined effects of transport and collision dynamics over long time horizons. When the output interval $\mathcal{T}_{\text{out}}$ lies far behind the input interval $\mathcal{T}_{\text{in}}$, local interactions are decorrelated and the evolution becomes increasingly global.


\begin{figure}[H]
\textbf{Physics-informed neural operator framework for the lattice Boltzmann equation}
\vspace{10pt}
\centering

\includegraphics[width=1\textwidth]{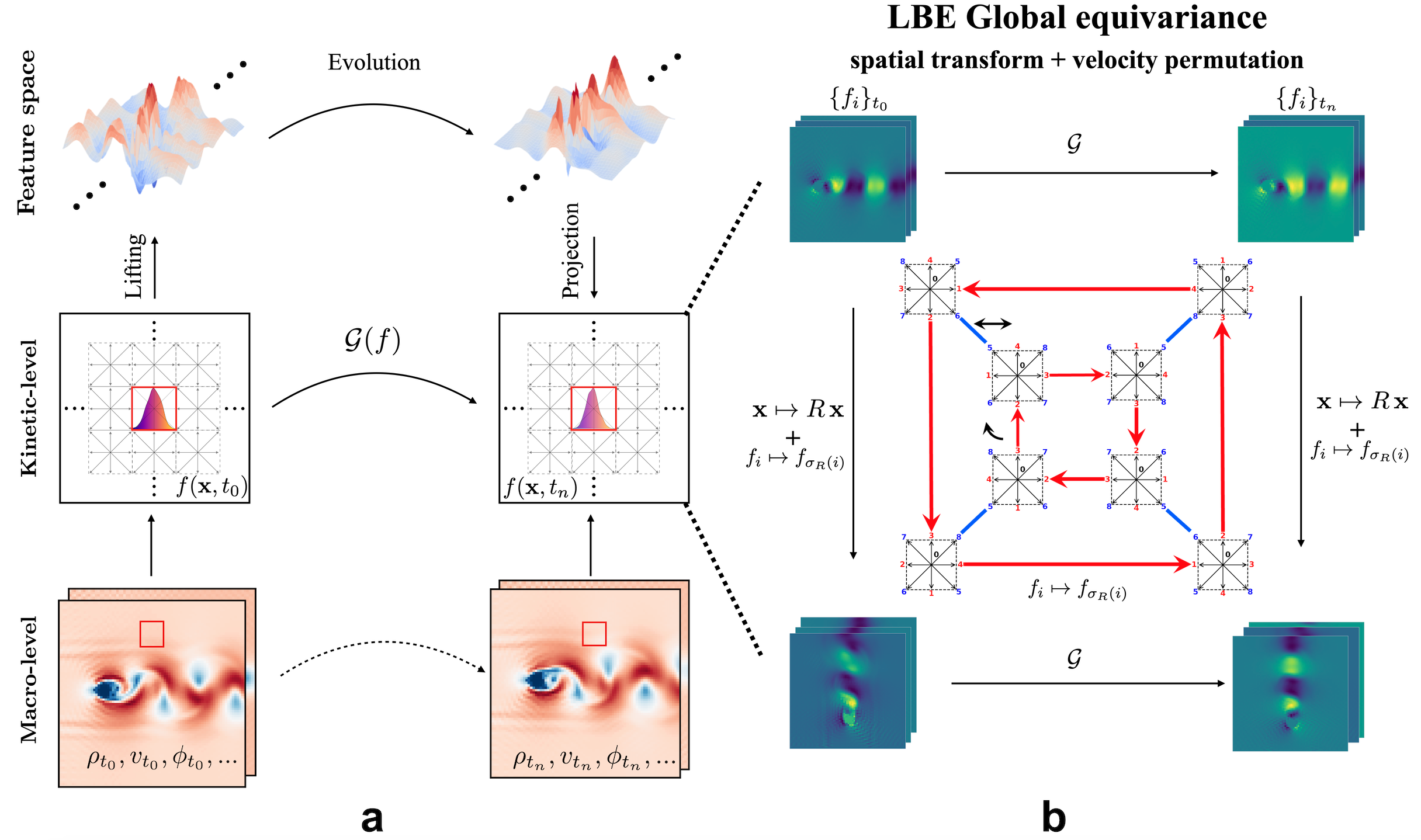}
\caption{\textbf{Physics-informed neural operator framework and global equivariance for the lattice Boltzmann equation.} Panel \textbf{a} \emph{Kinetic data-driven framework:} The operator $\mathcal{G}$ learns the long-time forward evolution of the distribution functions $\{f_i\}$. Starting from the kinetic-level fields $f(\mathbf{x},t_0)$, the input is lifted into a feature space, evolved by the neural operator, and projected back to obtain $f(\mathbf{x},t_n)$, where $t_n=t_0+n\Delta t$. Macroscopic observables such as density $\rho$, velocity $\mathbf{v}$, and phase field $\phi$ are recovered via moment maps (dotted arrows). Panel \textbf{b} \emph{Global equivariance under lattice symmetries:} For a group action $\mathbf{R}$ (rotation or reflection in $D_4$), the input field transforms as $[\mathbf{R} \cdot f]_i(\mathbf{x},t) = f_{\sigma_R(i)}(\mathbf{R}^{-1}\mathbf{x},t)$, combining a spatial transformation with permutation of velocity channels. The equivariance property $\mathcal{G}(\mathbf{R} \cdot f) = \mathbf{R} \cdot \mathcal{G}(f)$ ensures that predictions remain consistent with the discrete symmetries of the lattice Boltzmann formulation.}
\label{fig:sketch} 
\end{figure}

To approximate this forward operator, we employ physics-informed neural operators that generalize across spatial discretizations and enable inference on unseen grid resolutions. A key challenge in learning lattice Boltzmann dynamics is preserving the connection between mesoscopic distribution functions and the macroscopic physical quantities they encode, while also ensuring consistency with the underlying lattice structure of the system. To address this, we incorporate LBE-specific inductive biases into the training loss, including local moment-matching constraints to preserve meso-to-macroscale consistency and global equivariance of the distribution function fields under the action of the discrete symmetry group induced by the underlying local lattice structure. These constraints guide the learned operator to remain physically consistent across scales and symmetries.
Figure~\ref{fig:sketch}\textbf{a} provides an overview of the kinetic data-driven framework.  We begin by generating kinetic data through lattice Boltzmann simulations. The distribution fields $f(\mathbf{x},t_0)$ are lifted into a feature space, evolved through the neural operator $\mathcal{G}$, and projected back to obtain $f(\mathbf{x},t_n)$. In this work, we mainly examine the U-shaped Neural Operator~\cite{rahman2022u}, which implicitly learns this evolution via a data-driven composition of local and global operators, yielding approximate solutions at the kinetic level. From the predicted distributions, macroscopic observables such as density $\rho$, velocity $\mathbf{v}$, and phase field $\phi$ are recovered through moment maps (see~\cref{sec:method-lbm}). A detailed description of the unit conversion from lattice units (LU) to physical units is provided in Supplementary Information S2. 

Figure~\ref{fig:sketch}\textbf{b} illustrates the role of discrete symmetry groups (e.g., the dihedral group $D_4$ in D2Q9) in enhancing global equivariance, which requires applying both a spatial transformation $\mathbf{R}:\mathbf{x}\mapsto \mathbf{R}\mathbf{x}$ and a velocity-channel permutation $f_i \mapsto f_{\sigma_R(i)}$.  This property is formally expressed as $\mathcal{G}(\mathbf{R}\cdot f) = \mathbf{R}\cdot \mathcal{G}(f)$ and ensures that the learned operator respects the discrete symmetries of the lattice Boltzmann formulation (see Methods C and Supplementary Information S1 for the full proof). 
Training is performed with two loss formulations: a standard mean squared error (MSE) loss for data fidelity, and a physics-informed variant (MSE+Phys) that augments MSE with local moment-matching and global equivariance constraints (see~\cref{eq:total_loss}). 
Details of the network architecture are given in Supplementary Information S4.

\begin{table}[htbp]
  \centering
  \setlength{\tabcolsep}{3pt}
  \renewcommand{\arraystretch}{1.1}

  \rotatebox{90}{%
    \resizebox{0.86\textheight}{!}{%
      {\footnotesize
      \begin{tabular}{|c|c|c|c|c|c|}
        \hline
        \textbf{Physical Problems} & \textbf{Lattice Models} &
        \textbf{Training Dataset} & \textbf{Test Model Output} &
        \textbf{Input Snapshot} & \textbf{Output Snapshot} \\
        \hline

        \parbox[t]{2.4cm}{\centering von K\'arm\'an\\ vortex streets} &
        D3Q19 &
        \parbox[t]{4.2cm}{\raggedright
          $Re=180$, $R_c=3$ LU; prediction jumping time $50\,\Delta t$;\\
          $N_{\text{train}}=1600$
        } &
        \parbox[t]{4.2cm}{\raggedright
          $G(x,t_{\text{in}})\!\to\! f(\mathbf{x},t_{\text{out}})$;\\
          $t_{\text{in}}\!\in\![50\,\Delta t_{\text{coll}}\!:\!250\,\Delta t_{\text{coll}}]$;\\
          $N_{\text{val}}=200$, $N_{\text{test}}=200$
        } &
        \parbox[t]{3.2cm}{\centering
          \includegraphics[width=\linewidth]{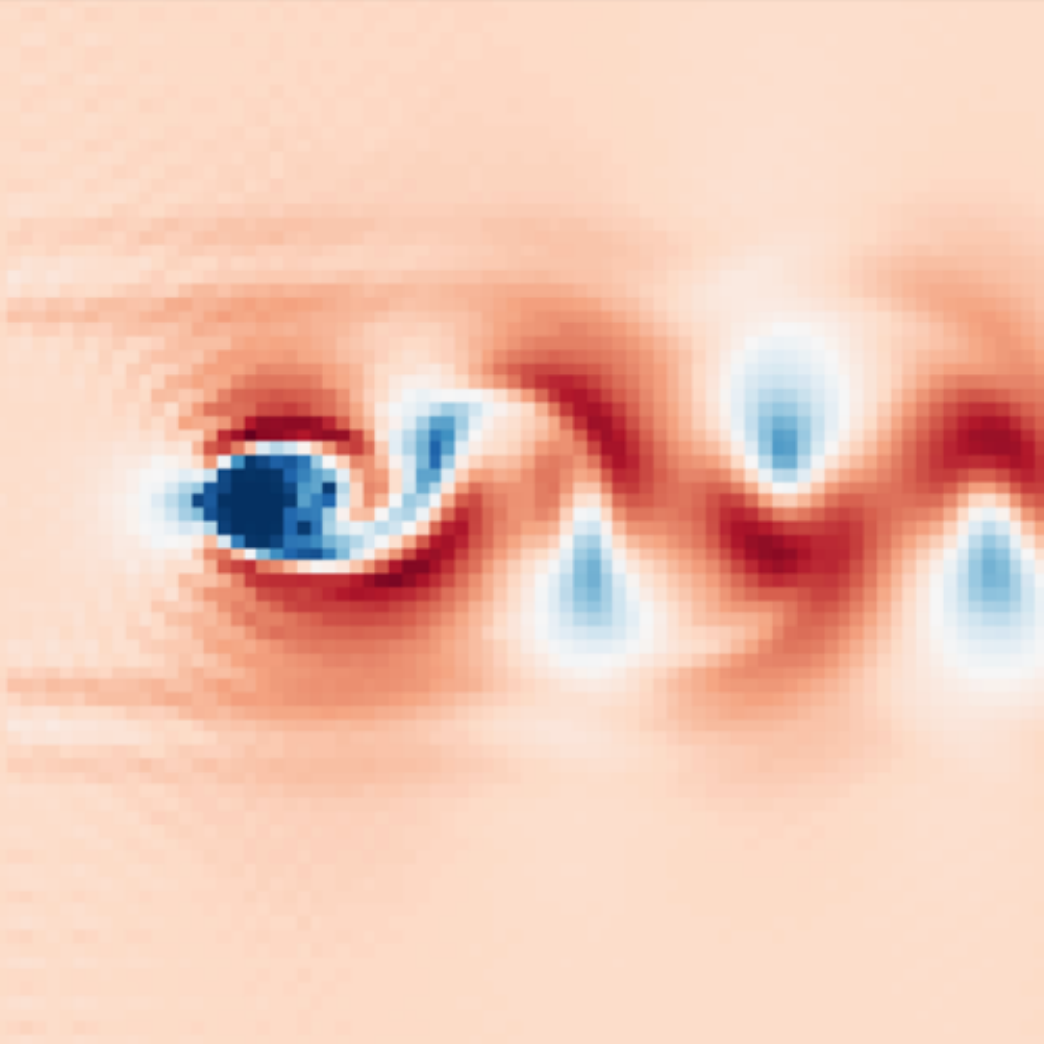}
        } &
        \parbox[t]{3.2cm}{\centering
          \includegraphics[width=\linewidth]{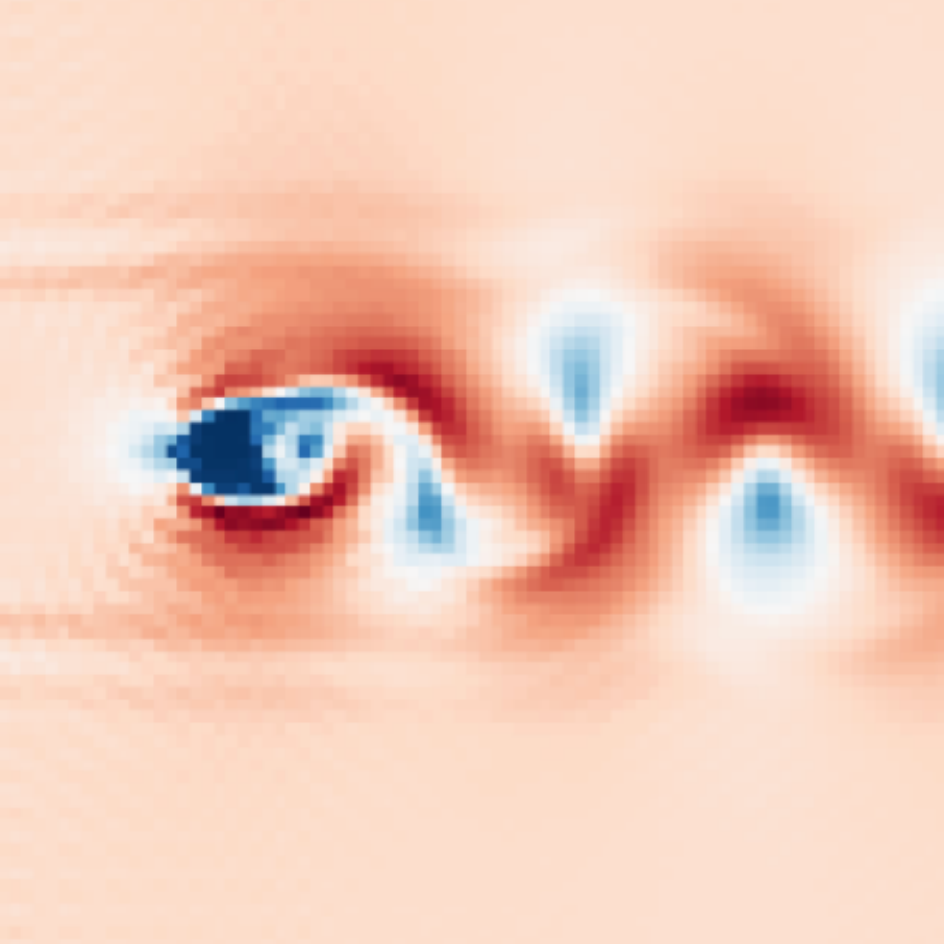}
        } \\
        \hline

        \parbox[t]{2.4cm}{\centering Ligament breakup} &
        D3Q19 &
        \parbox[t]{4.2cm}{\raggedright
          $x_c\!\in\![14\!:\!2\!:\!34],\ \phi_s\!\in\![0\!:\!0.04\!:\!2)$;\\
          prediction jumping time $10\,\Delta t$; $N_{\text{train}}=352$
        } &
        \parbox[t]{4.2cm}{\raggedright
          $G(x,t_0\!=\!0)\!\to\! f(\mathbf{x},t^*)$; $t^*\!\in\![0\!:\!1]$;\\
          $N_{\text{val}}=44$, $N_{\text{test}}=44$
        } &
        \parbox[t]{3.2cm}{\centering
          \includegraphics[width=.9\linewidth]{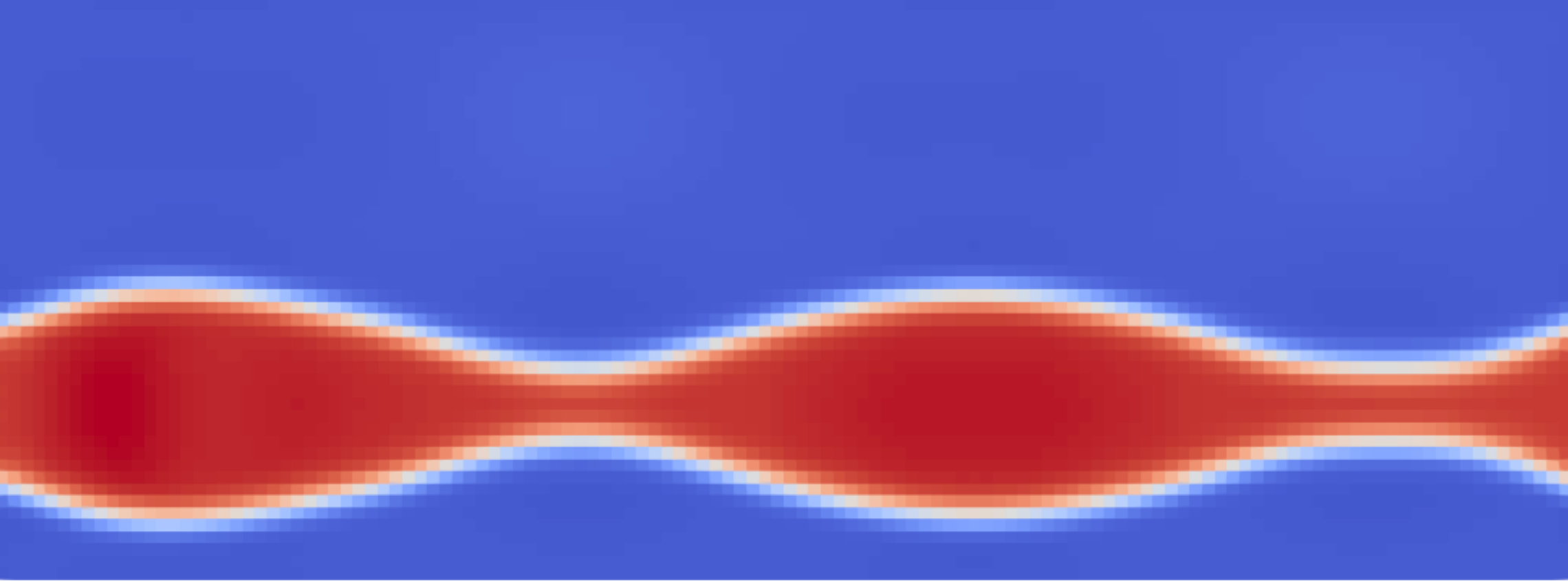}
        } &
        \parbox[t]{3.2cm}{\centering
          \includegraphics[width=.9\linewidth]{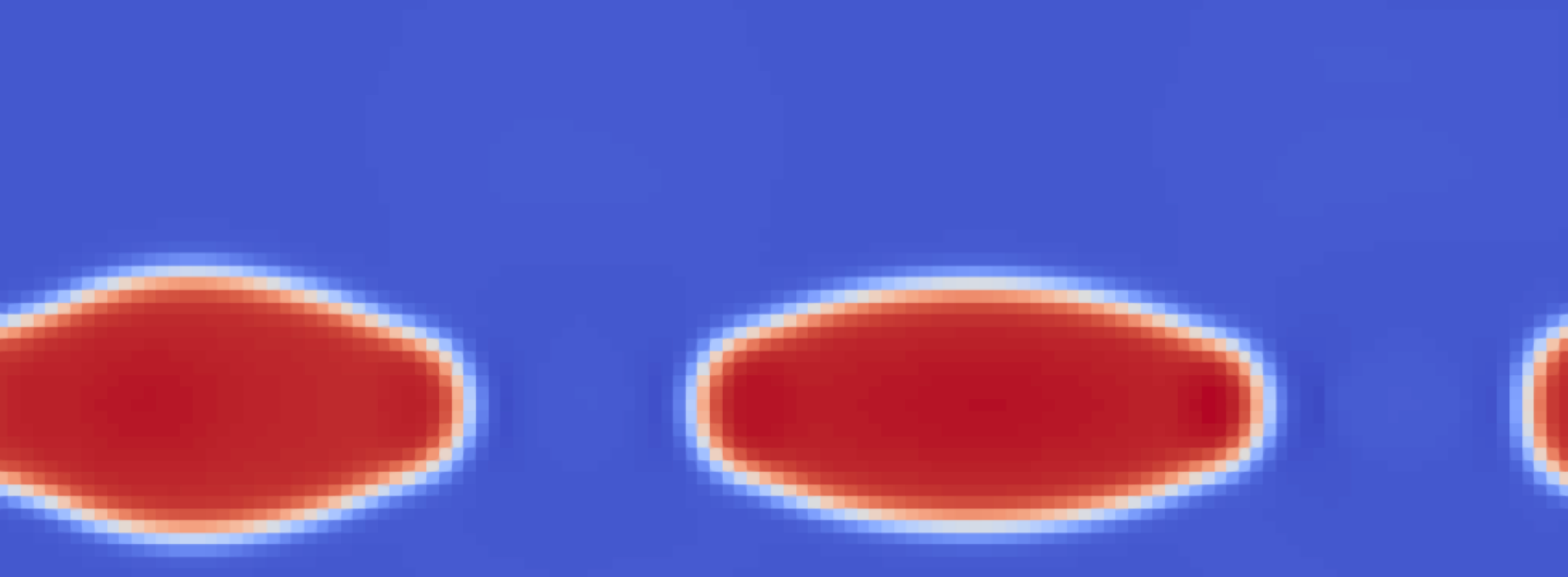}
        } \\
        \hline

        \parbox[t]{2.4cm}{\centering Bubble adhesion} &
        D2Q9 &
        \parbox[t]{4.2cm}{\raggedright
          $x_d\!\in\![26\!:\!1\!:\!56],\ y_d\!\in\![30\!:\!1\!:\!46)$;\\
          prediction jumping time $300\,\Delta t$; $N_{\text{train}}=396$
        } &
        \parbox[t]{4.2cm}{\raggedright
          $G(x,t_0\!=\!0)\!\to\! f(\mathbf{x},t^*)$; $t^*\!\in\![0\!:\!1]$;\\
          $N_{\text{val}}=50$, $N_{\text{test}}=50$
        } &
        \parbox[t]{3.2cm}{\centering
          \includegraphics[width=\linewidth]{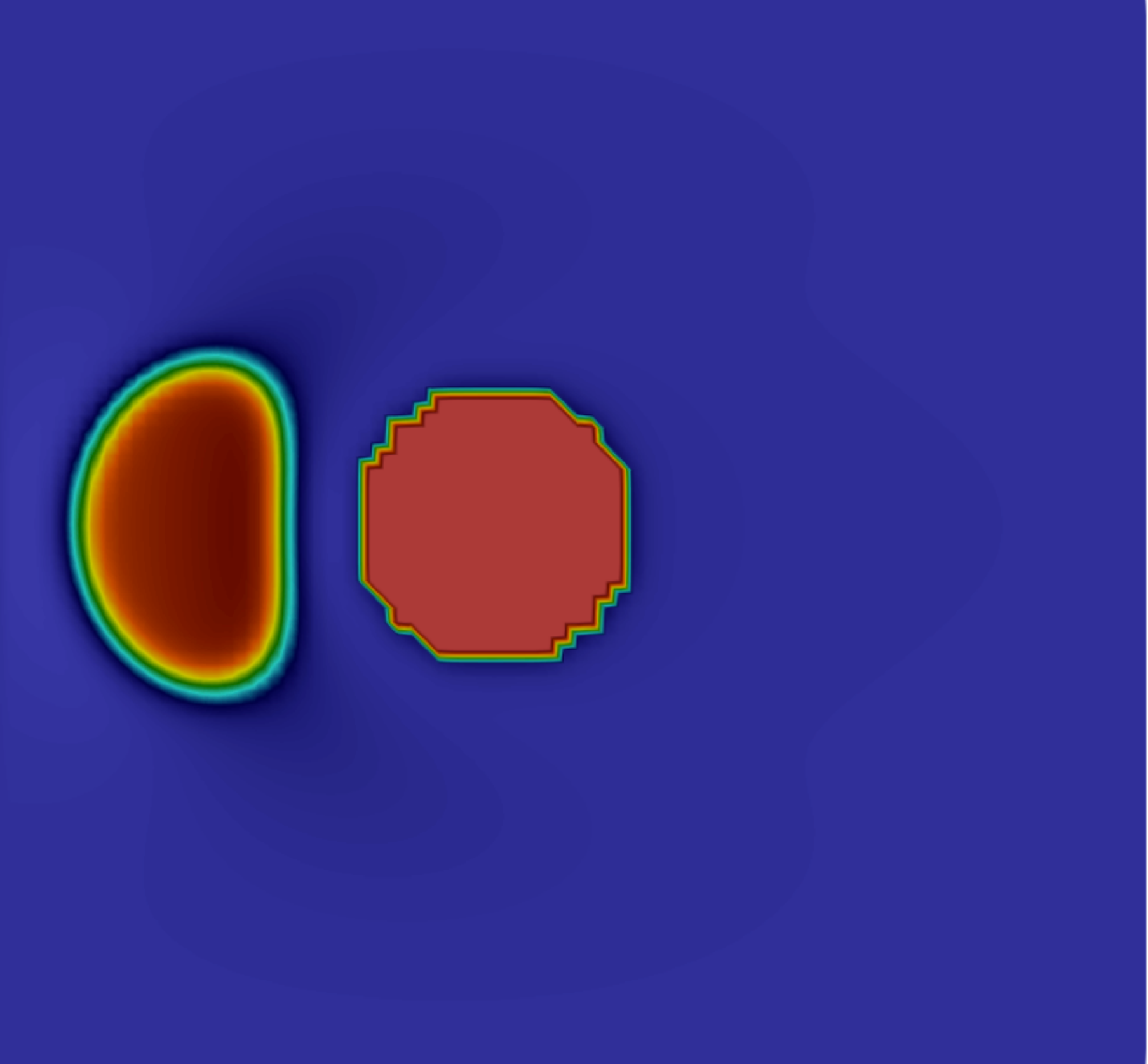}
        } &
        \parbox[t]{3.2cm}{\centering
          \includegraphics[width=\linewidth]{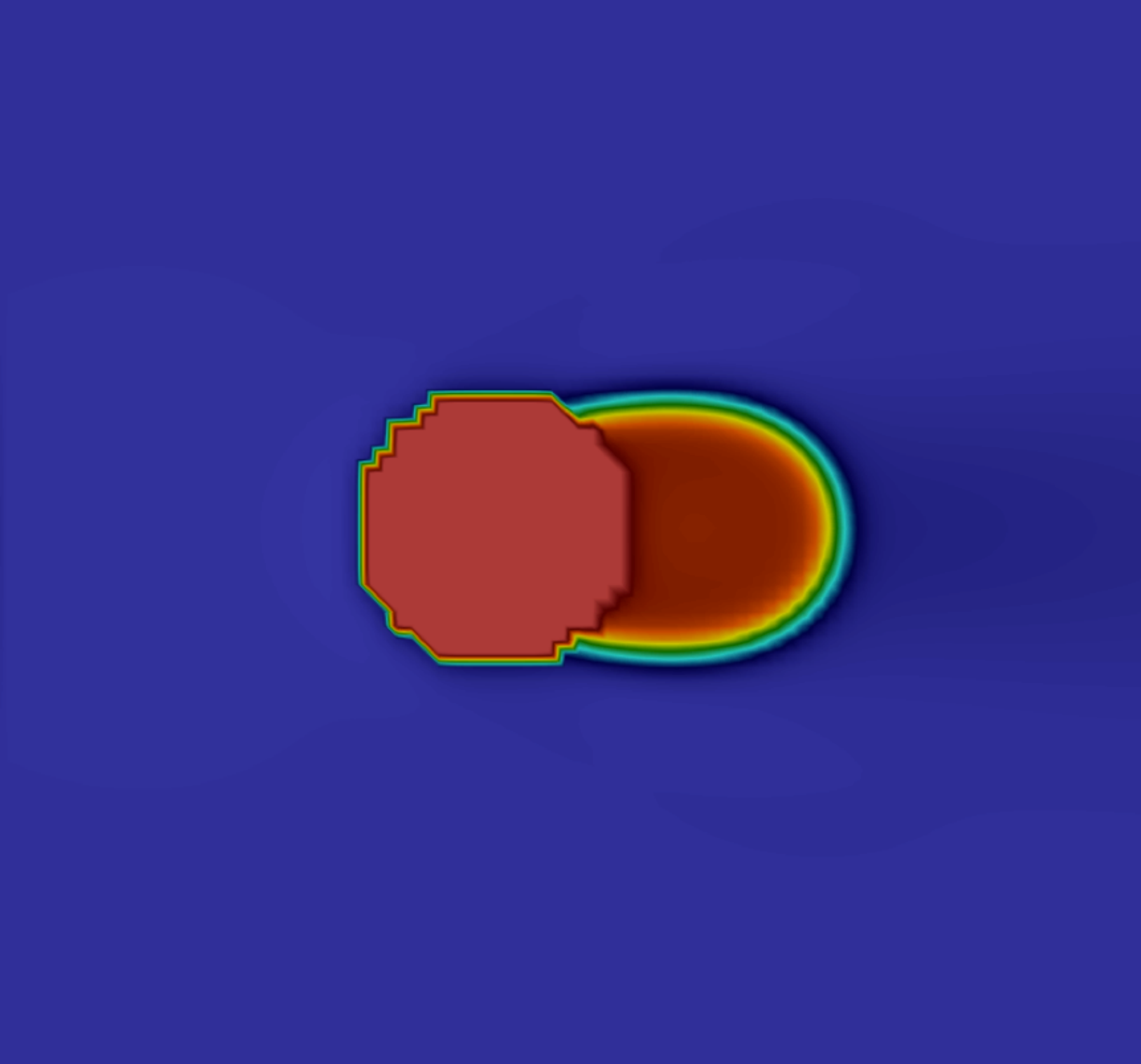}
        } \\
        \hline
      \end{tabular}
      }
    }
  }

  \caption{Schematic overview of the kinetic data-driven examples.}
  \label{tab:domain_figures}
\end{table}

\section{Applications}
We have assessed our proposed PINO for LBE framework with several applications of increasing complexity, including different physical phenomena governed by various LB models in 2D and 3D. In doing so, our framework has established fundamental principles for handling kinetic models that can be applied to general hydrodynamics set-ups, ensuring its broad applicability and reliability. Our evaluation begins with the von K\'arm\'an vortex street problem and progresses to more complex scenarios, including the breakup of a multiphase liquid cylinder and bubble adhesion on a solid cylinder. A systematic overview of these problems, including visualisations of the physical input–output mappings, is presented in Table~\ref{tab:domain_figures}. To validate the robustness of our framework, we also perform an ablation study by replacing the UNO backbone with FNO. Additional training details and extended results are provided in the Supplementary Information. The kinetic dataset is available from Xue and Coveney~\cite{xue_data}.
\subsection{Von K\'arm\'an vortex streets}
\begin{figure}[h]
\rotatebox{90}{
\begin{minipage}{\textheight}
\centering
\textbf{Learning von K\'arm\'an vortex streets from kinetic data}
\includegraphics[width=0.95\textwidth]{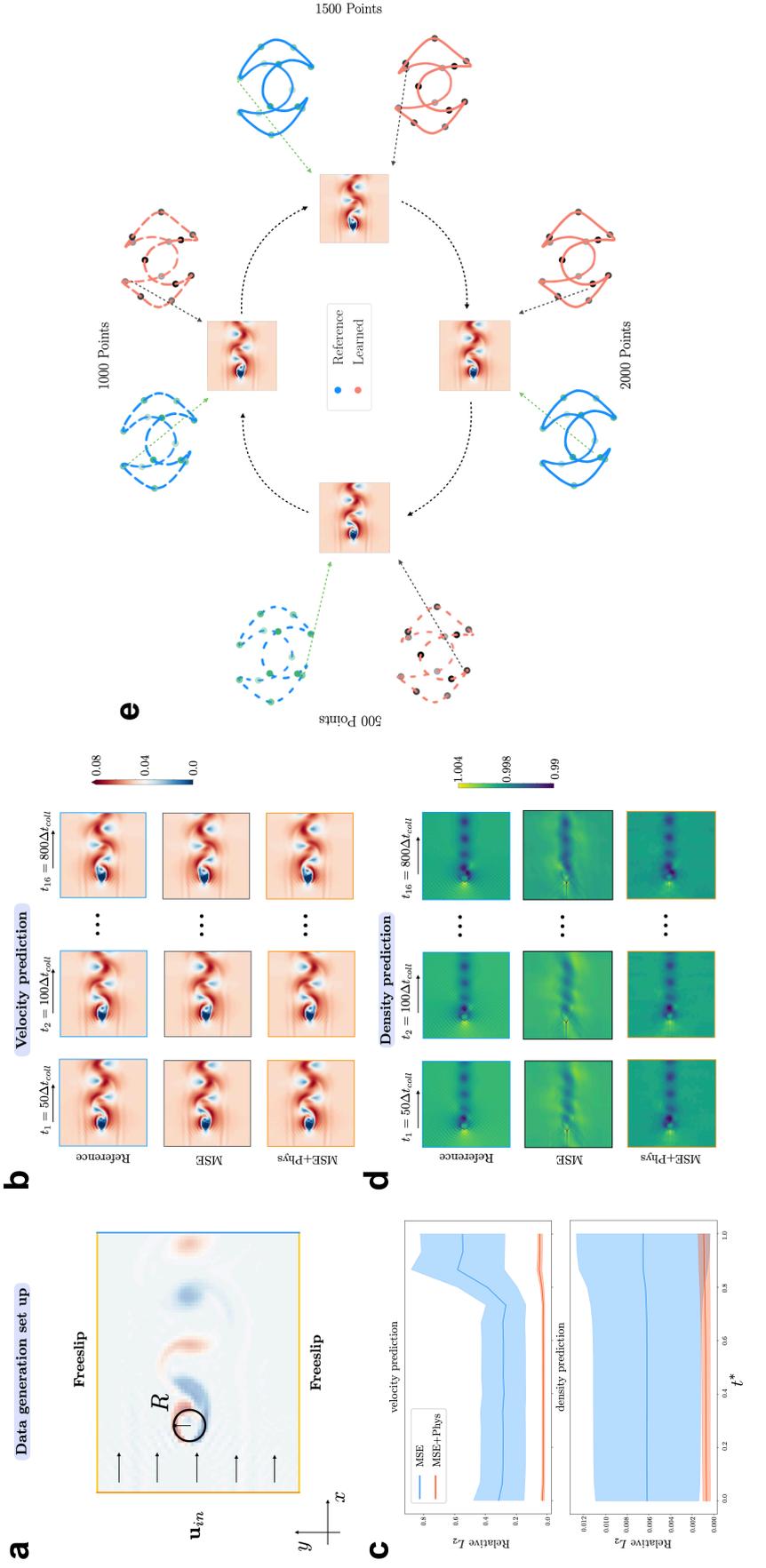}
\caption{Von K\'arm\'an vortex streets prediction using the kinetic data. The initial condition is randomly selected from a test dataset, and the prediction is generated autoregressively afterwards. Panel \textbf{a} demonstrates the sketch of the experimental setup. Panel \textbf{b} top row is the ground truth of the velocity field evolution. The middle row shows the prediction with the MSE case. The bottom row is the prediction with the MSE+Phys case. Panel \textbf{c} presents the relative $L_2$ evaluation for predicting macroscopic velocity and density over dimensionless time $t^*$. Panel \textbf{d} demonstrates the density evaluation over time. The top row is the ground truth of the density field evolution. The middle row shows the prediction based solely on the MSE constraint. The bottom row is the MSE+Phys constraint. Panel \textbf{e} shows the embedding orbits from kinetic data, comparing ground truth (blue) and learned embeddings (orange) across sample points from 500 to 2000 in time. Black and green dots highlight a snapshot of the von K\'arm\'an vortex street, illustrated by the vorticity plot with a blue-orange colour map.}
\label{fig:karman}
\end{minipage}
}
\end{figure}

As illustrated in~\cref{fig:karman} \textbf{a}, the 2D domain incorporates several boundary conditions. At the inlet, we impose a uniform velocity profile of $u_{in} = 0.01$ LU along the 
$x$ direction, where LU denotes lattice units. The outlet is defined by a pressure-free boundary condition~\cite{latt2008straight,xue2022synthetic,xue2023wall}. Additionally, freeslip boundary conditions are applied to both the top and bottom sides to eliminate drag from the boundary in the simulation~\cite{xue2022synthetic}. The radius of the cylinder is set to $R_c = 3$ LU. The Reynolds number, defined in the lattice Boltzmann framework as $Re = \frac{\mathbf{u}_{in} R_c}{\nu}$, with $\nu$ the kinematic viscosity in LU, is set to $Re = 180$. The cylinder boundary equips a no-slip boundary condition. The first row in Table~\ref{tab:domain_figures} shows the dataset configuration and the model output. The purpose of this example is to demonstrate that our framework can effectively learn from the kinetic dataset governed by~\cref{eq:lbe}, predict the evolution of the kinetic data without explicitly learning the collision kernel and perform self-regressive rollouts. In Fig.~\ref{fig:karman} \textbf{b}, the top row presents the ground truth evolution in the physical space evolution $t_1=50\Delta t$ up to $t_{16}=800\Delta t$ derived from the kinetic data follow~\cref{eq:momentum}, where $\Delta t$ represents a single collision time step. Each prediction step is $\Delta t_n = 50\Delta t$. Both the MSE constraint and the MSE+Phys constraint successfully captured the velocity evolution. 

We adopted an 80\%-10\%-10\% split for training, validation, and testing, respectively. To assess the fidelity of our ML models~\cite{coveney2024artificial,coveney2024sharkovskii,klower2023periodic}, the framework was trained independently 10 times and evaluated on a held-out test set of 200 samples (See Supplementary Information S5 and S6). In Fig.~\ref{fig:karman} \textbf{c}, we present the relative $L_2$ error in macroscopic predictions obtained using the MSE and MSE+Phys loss functions. The error metrics of relative $L_2$ is described in Supplementary Information S8. The plots show the evolution the error over normalized time $t^*=t/t_{end}$ with $t_{end}=6000\Delta t$ for velocity (top) and density (bottom), comparing models trained with standard MSE loss (blue) and physics-informed MSE+Phys loss (red).  Shaded regions indicate the standard deviation over 10 independently trained models. The MSE+Phys constraint consistently outperforms MSE alone, producing both lower mean error and reduced variance across the prediction horizon. The gap becomes more pronounced at later stages of the simulation, particularly for velocity, highlighting the stabilizing effect of the physics-based constraints over long-term rollouts. Figure~\ref{fig:karman} \textbf{d} shows the density evolution obtained through the kinetic data following Eq.~\ref{eq:density}. Due to the weak compressibility of LBE~\cite{succi2001lattice}, the predicted field captures minor density oscillations with the help of the MSE+Phys constraint, indicated by dark blue circles, which align well with the ground truth. However, the MSE constraint alone fails to capture the density variation in the tail flow after the cylinder. 

It is important to note that all results presented in this paper are in dimensionless units, which can be easily converted to physical units as described in Supplementary Information S2. The error evaluation for individual snapshots can be found in Supplementary Information S6. To better interpret the high-dimensional feature space, we employ t-SNE (t-distributed Stochastic Neighbour Embedding)~\cite{van2008visualizing}, an unsupervised dimensionality reduction technique, to project the feature space data into three-dimensional points. The results presented in Fig.~\ref{fig:karman} \textbf{e} use blue to represent the reference data and red to denote the learned data. As data points increase from 500 to 2000, these points coalesce into blue and red orbits, forming distinct embedding orbits for the reference and learned data. These closed orbits reveal the periodicity of the von K\'arm\'an vortex street. The close alignment between the learned and reference orbits indicates that the learned data successfully replicates the underlying dynamics of the ground truth at the kinetic level. To investigate this concept further, we examine the first period of von K\'arm\'an vortex streets. Thirteen sparse black and green dots mark 13 states within a single period of von K\'arm\'an vortex streets. Four of these representative states are displayed using vorticity plots, obtained by reconstructing the kinetic data into macroscopic velocity fields. When the orbit is densely populated with data points, it suggests that populating the embedding space with a large number of samples enables the neural network to accurately capture next-step predictions, thereby overcoming the limitations of single-step forward modelling. 

Our framework accelerates prediction from LB simulation by $\approx 363\times$ (see Supplementary Information Section~S3 for details). Importantly, we also tested a higher resolution of similar von K\'arm\'an vortex street dynamics at a size of $160\times 160$ LU. The reconstructed macroscopic velocity results agree well with the ground truth. Detailed information is presented in Supplementary Information S6. An ablation study comparing FNO models trained with and without physics-based constraints was also conducted. The model trained with the combined MSE+Phys loss consistently outperforms the one trained with MSE alone (see Supplementary Information, Section S7.1).




\subsection{Liquid cylinder breakup}
\begin{figure}[h]
\rotatebox{90}{
\begin{minipage}{\textheight}
\centering
\textbf{Learning ligament breakup from kinetic data}
\includegraphics[width=0.8\textheight]{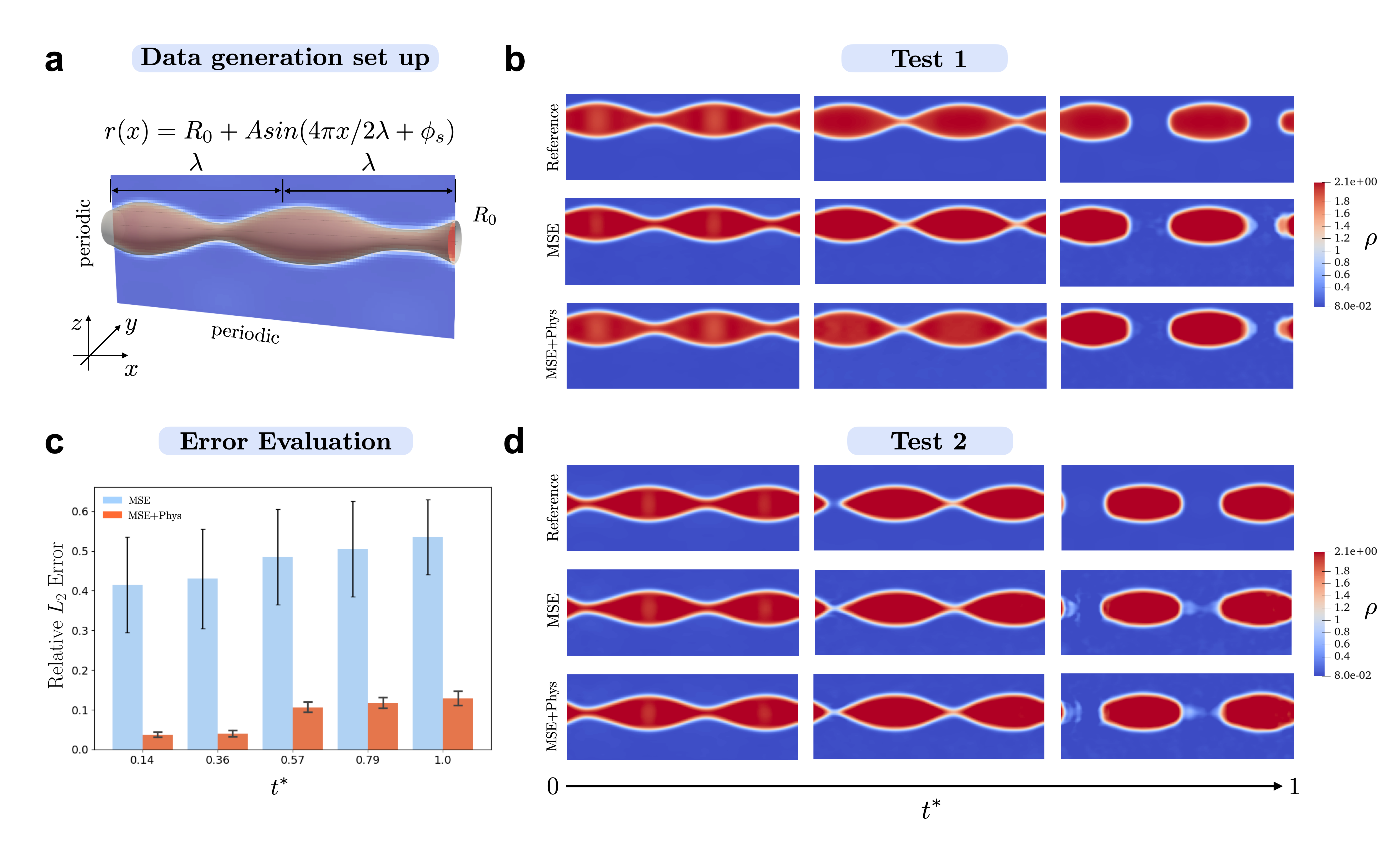}
\caption{Study of liquid ligament breakup. The test case is selected from a random initial configuration in the test dataset. Panel \textbf{a} illustrates the simulation setup for the dataset, showing the ligament radius $R_0$ and phase shift $\phi_s$. The wavelength of the simulation is denoted as $\lambda$. Panel \textbf{b} presents test case 1 results. The top row displays the ground truth of the ligament breakup process, the middle row shows the predicted breakup process learned solely using the MSE constraint, and the bottom row presents the prediction using MSE+Phys constraint. Panel \textbf{c} shows the error evaluation $L_2$ as a function of $t^*$, where $L_2$ is measured based on the reconstructed density using predicted kinetic data. Panel \textbf{d} presents test case 2; the top, middle, and bottom rows display the ground truth, the prediction using only the MSE loss, and the prediction using MSE+Phys loss, respectively.}
\label{fig:ligament} 
\end{minipage}%
}
\end{figure}

This example illustrates the capability of our learning framework to describe multi-phase problems. The cylinder ligament is set up within the 3-dimensional domain with periodic boundary conditions in the $x$, $y$ and $z$ directions. The ligament surface is modelled via $r(x)=R_0+A sin(4\pi x/2\Lambda + \phi_s)$, with $r(x)$ as the local radius of the cylinder, $\Lambda$ is the wavelength, $\phi_s$ is the phase shift, $A$ represents the amplitude of the sin function (Fig. ~\ref{fig:ligament}\textbf{a}). We set $A=2.5$ LU, while the centre of the ligament is located at $x_c$. The dataset is generated by varying $x_c\in[14:2:34]$ and $\phi_s\in[0.04:0.05:2)$, see in Table~\ref{tab:domain_figures} the second row. For training purposes, we take the $xz$ plane to be located in the middle cross section of the $y$ direction to focus on the region of interest. The prediction time interval is set to $\Delta t_n = 10 \Delta t$ to capture the rapid transitions occurring during ligament breakup. We divide the dataset into a split of 80\% training, 10\% validation, and 10\% testing for the training and evaluation process. The detailed data generation parameters for the simulations in~\cref{eq:lbe} can be found in Supplementary Information S5.

Figure~\ref{fig:ligament} \textbf{b} demonstrates a ligament breakup test case with dimensionless number $t^*=t/t_{end}$ starting from $t^* \in  \left [ 0, 1 \right ]$. Both the MSE constraint (middle row) and the MSE+Phys constraint (bottom row) are able to capture the breakup process. The MSE+Phys constraint prediction shows less spur currency near the breakup regime. In Fig.~\ref{fig:ligament} \textbf{d}, we further examine test case 2, which is also an unseen scenario in the test dataset, and we observe a similar situation for the MSE+Phys constraint which is marginally better than solely using the MSE constraint. 

Although the MSE+Phys constraint initially appears to be only slightly more effective than the MSE constraint alone, a more detailed comparison reveals a significant advantage. To rigorously assess this, we conducted 10 independent training models for the constraints of MSE and MSE+Phys (see Supplementary Information S6). Our analysis shows that the MSE constraint tends to converge prematurely with a high $L_2$ error, often getting stuck in local minima, which leads to prediction failures. In contrast, the MSE+Phys constraint consistently enhances the training success rate, yielding more robust and reliable outcomes. Further testing on a set of 10 pre-trained models corroborates these findings. Figure~\ref{fig:ligament} \textbf{c} shows the relative $L_2$ error of macroscopic predictions evaluated at five normalized time points $t^* = {0.14,\ 0.36,\ 0.57,\ 0.79,\ 1.0}$, comparing models trained with standard MSE loss (blue bars) and physics-informed MSE+Phys loss (orange bars). Error bars represent 95\% confidence intervals across 10 independently trained models. The MSE+Phys constraint consistently reduces both the mean error and variance across all time points, with the performance gap increasing over time. This result highlights the improved stability and predictive accuracy achieved by incorporating the physics-informed loss during training. The ablation study for FNO follows similar trends for both constraints (See Supplementary Information section S7.2). Additionally, The model achieves a $\approx 14\times$ speedup relative to the corresponding LB simulation (details in Supplementary Information Section~S3).

\subsection{Bubble adhesion}

We demonstrate our learning framework applied to the phase-field kinetic model coupled with the hydrodynamic kinetic model for simulating bubble adhesion in the presence of a cylinder. In this example, we learn the phase-field LB model on 2D, which equivariently matches the Cahn-Hilliard equations~\cite{fakhari2010phase} (see the supplementary information S2.). Figure~\ref{fig:binary} \textbf{a} illustrates the general simulation setup. The simulation data generation is conducted within a 2D domain of size $L_x \times L_y$, measuring \RE{$84 \times 78$} LU. This domain comprises three phases: $\phi_1$, $\phi_2$, and $\phi_3$, representing the bubble, fluid field, and solid cylinder, respectively. The fluid enters the domain with an inlet velocity directed as shown by the arrows, while the top and bottom boundaries are defined by free-slip boundary conditions. The outlet boundary is set as a pressure-free condition. The cylinder’s centre coordinates, $x_d$ and $y_d$, are uniformly distributed across the domain, resulting in a total dataset size of $N_{\text{total}} = 496$, as indicated in the third row of Table~\ref{tab:domain_figures}. Each prediction time interval is set to $\Delta t_n = 300\Delta t$. We split the dataset into 80\% training, 10\% validation, and 10\% testing. Specifically, we used $N_{\text{train}} = 396$ samples for training, $N_{\text{val}} = 50$ for validation, and $N_{\text{test}} = 50$ for testing. The training and validation sets are used to optimize model performance, while the test set includes previously unseen configurations. Detailed data generation parameters related to the simulations governed by~\cref{eq:lbe} are provided in Supplementary Information S5.

\begin{figure}[h]
\rotatebox{90}{
\begin{minipage}{\textheight}
\textbf{Learning bubble adhesion}
\centering

\includegraphics[width=0.7\textheight]{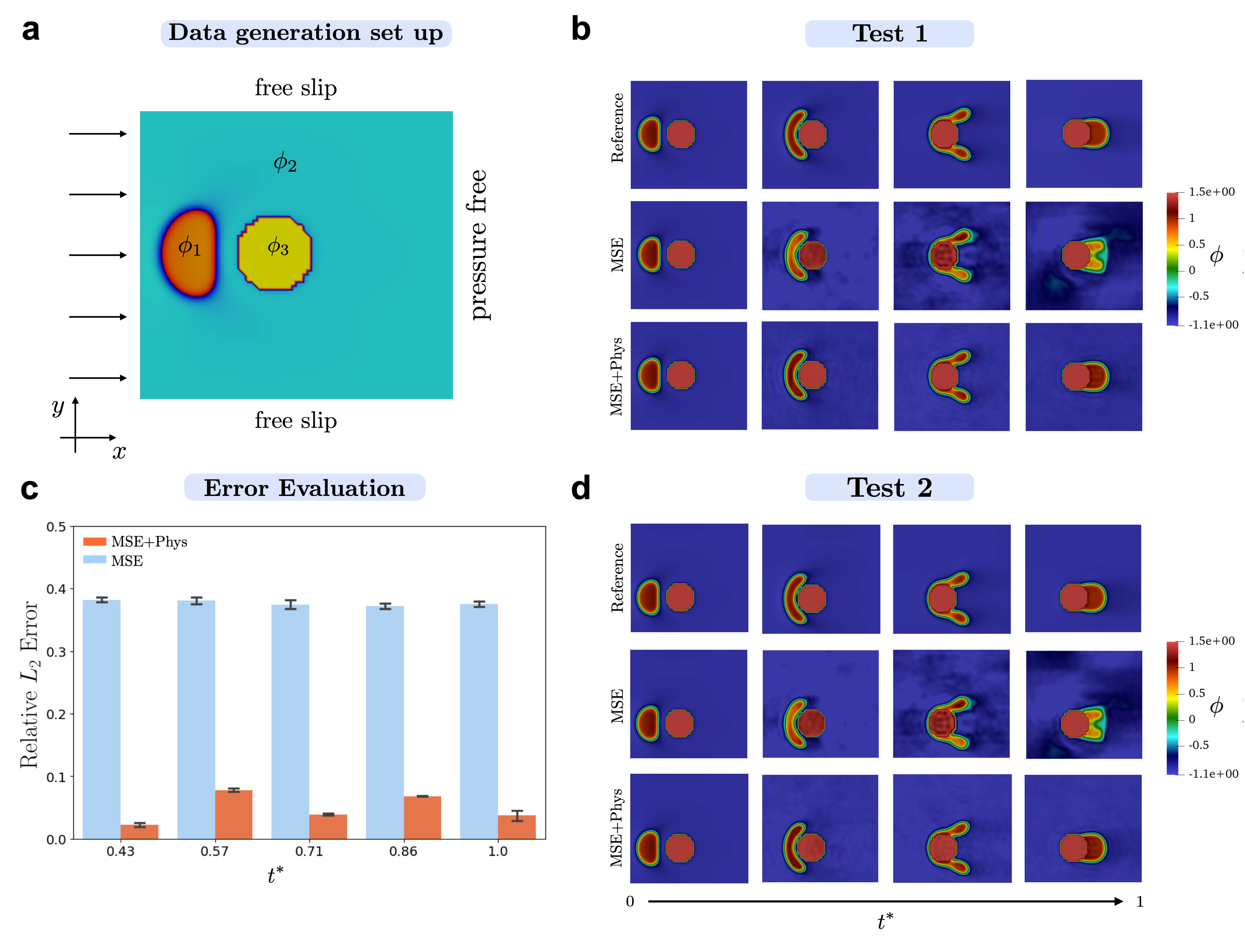}
\label{fig:capture}
\vspace{-15pt}
\caption{Binary phase capture by a cylinder in 2D. Panel \textbf{a} presents a sketch of the simulation setup, showing three phases: fluid 1, $\phi_1$, in orange, fluid 2, $\phi_2$, in green, and the solid cylinder, $\phi_3$, in yellow. Panel \textbf{b} depicts the evolution of a bubble over a sphere with radius $R_s = 4$ LU. The top row shows the ground truth phase field evolution, the middle row presents the phase field prediction using the MSE constraint, and the bottom row illustrates the prediction using MSE+Phys constraints, as described in Eq.~\ref{eq:total_loss}. Panel \textbf{c} shows the error evaluation $L_2$ as a function of dimensionless $t^*$, where $L_2$ is measured based on the reconstructed phase using predicted kinetic data. Panel \textbf{d} presents test case 2 over time, with the top, middle, and bottom rows displaying the ground truth, the prediction using the MSE loss, and the prediction using the MSE+Phys loss, respectively.}
\label{fig:binary} 
\end{minipage}
}
\end{figure}
Figure~\ref{fig:binary} \textbf{b} and \textbf{d} show two test cases for bubble adhesion at different locations. The bubble crosses over the cylinder, splits into two sub-bubbles and then merges again. The results show that the MSE+Phys constraint is able to capture the complete dynamics, whereas the MSE constraint can only capture the dynamics up to the splitting process, failing to merge the bubble afterwards. In general, the MSE+Phys constraint case outperforms the MSE-only case. Next, we conducted 10 independent training runs for both the MSE and MSE+Phys constraint configurations (see Supplementary Information S6 for details on the training progression and $L_2$ convergence). A comparative evaluation of the relative error is shown in Fig.~\ref{fig:binary} \textbf{c}, where the blue and orange bars correspond to models trained with MSE and MSE+Phys losses, respectively. The relative $L_2$ error is computed based on the macroscopic phase-field variable $\phi$ reconstructed from the underlying kinetic data. Error bars represent the 95\% confidence interval across the ensemble of trained models. As shown, the MSE+Phys case consistently yields substantially lower error across all time points, demonstrating improved predictive accuracy and robustness. Moreover, our approach provides a $\approx 609\times$ acceleration over the LB solver (see Supplementary Information Section~S3).

Finally, we conduct an ablation study on FNO for the bubble adhesion study, which reveals consistent trends across both types of constraints (see Supplementary Information, Section S7.3). This example showcases the versatility of our kinetic data-driven PINO method. Our approach can also learn phase-field-based kinetic models~\cite{fakhari2010phase}, demonstrating the generality of our framework.
\vspace{-\baselineskip}
\section{Conclusion}
This paper presents a physics-informed neural operator, implemented using the U-shaped neural operator architecture, to learn the fast-forward dynamics of the lattice Boltzmann equation at the kinetic level. UNO is chosen for its memory efficiency, making it well suited for large-scale kinetic modelling. Rather than relying on explicit collision kernel modelling or step-by-step integration, our approach learns the long-range evolution of the distribution functions, effectively bypassing the limitations imposed by the collision operator and enabling fast-forward prediction over extended time intervals. To ensure physical fidelity, we design a composite loss function that integrates mean squared error with local moment-matching and global equivariance constraints. The moment-matching terms promote accurate meso-to-macroscale consistency by preserving quantities such as density and momentum, while the equivariance constraint regularises symmetry under global transformations. Crucially, this global equivariance is induced by the underlying local lattice structure, ensuring that the learned dynamics remain consistent with the discrete symmetries inherent in the LBE formulation. We evaluate the proposed framework on three representative problems: von K\'arm\'an vortex streets, liquid cylinder breakup and  bubble adhesion. These examples span a wide range of kinetic regimes, including vortex shedding, multiphase interfacial dynamics and  phase-field modelling. By predicting the evolution of single-particle distribution functions, our model reconstructs macroscopic quantities such as velocity, density and  phase fields without requiring explicit macroscopic closures or predefined lattice rules. This physics-informed neural operator framework also offers extensibility to other physical observables, such as flux tensors, through higher-order moment-matching loss. Ablation studies replacing the U-shaped neural operator with the Fourier neural operator confirm that the proposed physics-informed constraints yield consistent performance improvements across architectures. Throughout all these applications, our framework achieves acceleration over conventional lattice Boltzmann solvers by between two and three orders of magnitude. The ability to capture complex, coupled physical dynamics—including small-scale oscillations and long-range transport—using fast-forward evolution of the distribution functions makes our framework a promising tool for data-driven modelling of physical systems governed by partial differential equations, while preserving their underlying kinetic structure.

\vspace{-\baselineskip}

\section{Methods} \label{Section: Methods}
\subsection{Physics-Informed training of neural operators for the lattice Boltzmann equation} \label{sec:method_no_lbe}

We approximate the global operator $\mathcal{G}$ introduced in Section~II using neural operator architectures. We employ either the U-shaped Neural Operator or the Fourier Neural Operator to parameterize the spatiotemporal mapping from $f_{\text{in}}$ to $\hat f_{\text{out}}$. Detailed formulations of both architectures are provided in Supplementary Information~S4. To ensure that the learned operator remains faithful to the underlying kinetic physics of the LBE, we train it with the physics-informed loss described below, which augments standard data fidelity with moment-matching and equivariance constraints.
To train the neural operator, we define a composite loss function that combines data fidelity with physics-informed constraints. The total loss is  
\begin{equation}\label{eq:total_loss}
\mathcal{L} \;=\; 
\lambda_{\text{MSE}}\,\mathcal{L}_{\text{MSE}}
\;+\; \lambda_{\text{mom},0}\,\mathcal{L}_{\text{mom},0}
\;+\; \lambda_{\text{mom},1}\,\mathcal{L}_{\text{mom},1}
\;+\; \lambda_{\text{equiv}}\,\mathcal{L}_{\text{equiv}},
\end{equation}
where $\mathcal{L}_{\text{MSE}}$ measures the discrepancy between predicted and reference distribution functions, $\mathcal{L}_{\text{mom},0}$ and $\mathcal{L}_{\text{mom},1}$ regularizes meso-to-macroscopic of zeroth- and first-order moments (mass and momentum), and $\mathcal{L}_{\text{equiv}}$ regularizes moment equivalence between predicted and reference distributions. The coefficients $\lambda_{\text{MSE}}, \lambda_{\text{mom},0}, \lambda_{\text{mom},1}, \lambda_{\text{equiv}}$ are non-negative coefficients that balance the contribution of each term. $\mathcal{L}_{\text{equiv}}$ penalizes violation of global LBE equivariance (See section Methods C). The moment matching losses are defined by:
\begin{align}
\mathcal{L}_{\text{mom},0} &= 
\left\| \sum_{i=0}^{Q-1} \hat{f}_i(\cdot, \cdot) -
       \sum_{i=0}^{Q-1} f_i(\cdot, \cdot) \right\|_2^2,  \label{eq:moment0} \\
\mathcal{L}_{\text{mom},1} &= 
\left\| \sum_{i=0}^{Q-1} \hat{f}_i(\cdot, \cdot)\,\mathbf{c}_i -
       \sum_{i=0}^{Q-1} f_i(\cdot, \cdot)\,\mathbf{c}_i \right\|_2^2,  \label{eq:moment1}
\end{align}
where $\hat{f}_i$ is the predicted distribution functions. These losses are computed locally across space and time to ensure the preservation of meso-to-macroscale consistency. Higher-order moments are not included, as the training data does not reliably resolve them, and they introduce sensitivity to numerical noise.

To ensure that the learned dynamics respect the global equivariance property of the lattice Boltzmann equation, we incorporate an equivariance loss that combines global spatial transformations with local permutations of the discrete velocity channels, both determined by the underlying lattice structure. Although local equivariance is commonly regularized in LBE learning frameworks~\cite{corbetta2023toward}, it becomes unreliable over long time intervals due to the compounded effects of streaming and collision, which break local symmetry even when initial conditions are symmetric. Since our neural operator models long-time evolution by directly predicting large temporal jumps rather than simulating intermediate steps, enforcing local equivariance is inappropriate. Instead, we promote global equivariance, defined as the invariance of the full distribution field under joint spatial transformation and velocity permutation, which remains physically meaningful over extended time horizons. We therefore define a global equivariance loss:

\begin{equation}
\mathcal{L}_{\text{equiv}} = \sum_{\mathbf{R} \in \mathcal{S}_h} \left\| \mathcal{G}(\mathbf{R} \cdot f) - \mathbf{R} \cdot \mathcal{G}(f) \right\|^2,
\end{equation}
where $\mathcal{S}_h$ denotes the discrete symmetry group of the lattice (e.g., dihedral group $D_4$ in 2D), and the group action $\mathbf{R} \cdot f$ applies both spatial transformations and velocity index permutations. This constraint promotes physically consistent evolution of the distribution functions under rigid transformations of the full system state. The weights $\lambda_{\text{MSE}}$, $\lambda_{\text{mom},0}$, $\lambda_{\text{mom},1}$, and $\lambda_{\text{equiv}}$ are treated as hyperparameters and selected based on validation performance. A detailed study of their setup is provided in Supplementary Information S3.

\subsection{The lattice Boltzmann method}\label{sec:method-lbm}

This work uses the LB simulation to generate the kinetic data. The detailed description of the lattice Boltzmann method is given in Supplementary Information S1. We use both two-dimensional and three-dimensional LB models, referred to as D2QX and D3QX, where $X$ denotes the number of discrete velocities for the lattice cell. The governing equation for the distribution function $f_i(\bx,t)$, accounting for collision and forcing, can be expressed as:
\begin{equation}
\label{eq:lbe}
f_i(\bx+\bc_i \Delta t, t+\Delta t) = f_i(\bx, t) + \Omega\!\left[f_i(\bx,t )-f_i^{\rm eq}(\bx,t )\right] + F_i(\bx,t),
\end{equation}
where $\Omega$ denotes the collision operator, $\bc_i$ represents the $i$th discrete velocity vector, $f_i^{\rm eq}$ is the equilibrium distribution function, and $F_i$ is the forcing term. A common choice is the Bhatnagar–Gross–Krook (BGK) collision operator, which can be written as $\Omega(\bx,t) = -\tfrac{1}{\tau}$. Other collision operators, for example, the multiple relaxation time (MRT) kernel~\cite{d2002multiple} and the cumulant kernel~\cite{geier2015cumulant}, can also be used for higher Reynolds number flows. A detailed description of the collision operators employed in this work is provided in Supplementary Information S1. 

Macroscale quantities such as density and momentum can be obtained through moment-matching as defined in~\cref{eq:moment0} and~\cref{eq:moment1}. Higher-order moments are not considered in this work, as the LBM data used does not preserve physically meaningful information beyond the first moment. The discretized expressions for the conserved macroscopic quantities are:
\begin{equation}\label{eq:density}
\rho(\bx, t) = \sum_{i=0}^{Q-1} f_i(\bx, t),
\end{equation}
\begin{equation}\label{eq:momentum}
\rho(\bx, t)\bu(\bx, t) = \sum_{i=0}^{Q-1} f_i(\bx, t)\bc_i,
\end{equation}
where $Q$ is the number of discrete velocity directions.

\subsection{Global equivariance induced by lattice symmetries}
The discrete symmetry groups $D_4$ (for 2D lattice models) and $O_h$ (for 3D lattice models) 
are not only essential to the physical structure of the lattice Boltzmann models but also play 
a central role in our neural operator framework. Specifically, these local lattice symmetries 
induce a global equivariance structure over the full distribution function fields. In the context 
of our model, global equivariance refers to the invariance of the neural operator's output under 
joint spatial transformations and velocity-index permutations, both defined by the corresponding 
lattice symmetry group. Enforcing global equivariance in this way ensures that the predicted 
evolution of the system remains consistent under rigid transformations of the entire domain, 
making the learned dynamics physically meaningful over long time jumps. 

Here, $\mathbf{R} \in \mathcal{S}_h$ denotes a group element acting on spatial coordinates and 
associated with a permutation $\sigma_{\mathbf{R}}$ on the velocity indices such that:
\begin{equation}
\mathbf{R}\,\mathbf{c}_{\sigma_{\mathbf{R}}(i)} = \mathbf{c}_i, \quad \forall i.
\end{equation}

Then, the group action on the distribution field is defined as:
\begin{equation}
[\mathbf{R} \cdot f]_i(\mathbf{x}, t) := f_{\sigma_{\mathbf{R}}(i)}(\mathbf{R}^{-1} \mathbf{x}, t).
\label{eq:group-action}
\end{equation}

We say that the neural operator $\mathcal{G}$ is globally equivariant if:
\begin{equation}
\mathcal{G}(\mathbf{R} \cdot f) = \mathbf{R} \cdot \mathcal{G}(f), \quad \forall \mathbf{R} \in \mathcal{S}_h.
\label{eq:global-equivariance}
\end{equation}

This relation ensures that a rigid transformation of the input field, rotating or reflecting the entire spatial domain and permuting velocity channels accordingly, results in a correspondingly 
transformed output field. Unlike local equivariance, which is defined at the level of individual lattice cells and may break due to streaming dynamics, global equivariance captures symmetries at 
the field level and remains robust over long time horizons. It is important to note that we regularize toward the {discrete symmetry group induced by the lattice, rather than the full continuous rotation group. For example, the D2Q9 lattice is invariant only under the dihedral group $D_4$ (rotations by multiples of $90^\circ$ and reflections), and the 
D3Q19 lattice under the cubic symmetry group $O_h$. Our equivariance loss therefore regularizes these discrete symmetries, which are exactly those respected by the underlying lattice Boltzmann formulation. The full proof of global equivariance for the lattice Boltzmann equation is provided in Supplementary Information S2.5.

   
\bibliographystyle{unsrtnat}
\bibliography{prex}

\section{Acknowledgements}
P.V.C. and X.X. acknowledge funding support from the European Commission CompBioMed Centre of Excellence (Grant No. 675451 and 823712). Support from the UK Engineering and Physical Sciences Research Council under the following projects ``UK Consortium on Mesoscale Engineering Sciences (UKCOMES)" (Grant No.
EP/R029598/1) and ``Software Environment for Actionable and VVUQ-evaluated Exascale Applications (SEAVEA)" (Grant No. EP/W007711/1) is gratefully acknowledged. P.V.C. and X.X. acknowledge 2024-2025 DOE INCITE award for computational resources on supercomputers at the Oak Ridge Leadership Computing Facility under the ``COMPBIO3" project. P.V.C. and X.X. acknowledge the use of resources provided by the Isambard-AI National AI Research Resource (AIRR). Isambard-AI is operated by the University of Bristol and is funded by the UK Government’s Department for Science, Innovation and Technology (DSIT) via UK Research and Innovation; and the Science and Technology Facilities Council [ST/AIRR/I-A-I/1023]. P.V.C and X.X. acknowledge Leibniz Supercomputing Centre (LRZ) for access to the BEAST GPU cluster. 

\section{Declaration of interests}
The authors declare that they have no known competing financial interests or personal relationships that could have appeared to influence the work reported in this article.

\section{Contributions}
X.Xue, S.Cheng, Y.Yang, X.Cheng, and P.V.Coveney conceived the initial plan for this research. X.Xue designed and conducted the numerical simulations. M.F.P. ten Eikelder, X.Xue, and X.Cheng accomplished the theoretical framework. X.Xue, M.Gao, S.Wang, Y.Yang performed machine learning training.  X.Xue, M.Gao, M.F.P. ten Eikelder, Y.Yang, Y.He, X.Cheng, and S.Wang performed the results analysis. P.V.Coveney, Y. Hu and S.Cheng provided supervision. P.V.Coveney acquired funding and provided access to computing resources.
\section{Competing of interests}
The authors declare that they have no competing interests.

\end{document}